\documentclass[sigconf]{acmart}

\AtBeginDocument{%
  \providecommand\BibTeX{{%
    \normalfont B\kern-0.5em{\scshape i\kern-0.25em b}\kern-0.8em\TeX}}}

\usepackage{todonotes}
\usepackage{enumitem}
\setlist[itemize]{leftmargin=*}

\setcopyright{rightsretained}

\begin{document}

\fancyhead{}

\copyrightyear{2019} 
\acmYear{2019} 
\acmConference[MM '19]{Proceedings of the 27th ACM International Conference on Multimedia}{October 21--25, 2019}{Nice, France}
\acmBooktitle{Proceedings of the 27th ACM International Conference on Multimedia (MM '19), October 21--25, 2019, Nice, France}
\acmPrice{15.00}
\acmDOI{10.1145/3343031.3350889}
\acmISBN{978-1-4503-6889-6/19/10}

\title{Who, Where, and What to Wear? Extracting Fashion Knowledge from Social Media}


 \author{Yunshan Ma}
 \affiliation{
     \institution{National University of Singapore}
 }
 \email{yunshan.ma@u.nus.edu}

 \author{Xun Yang}
 \authornote{Corresponding author.}
 \affiliation{
     \institution{National University of Singapore}
 }
 \email{xunyang@nus.edu.sg}

 \author{Lizi Liao}
 \affiliation{
     \institution{National University of Singapore}
 }
 \email{liaolizi.llz@gmail.com}

 \author{Yixin Cao}
 \affiliation{
     \institution{National University of Singapore}
 }
 \email{caoyixin2011@gmail.com}

 \author{Tat-Seng Chua}
 \affiliation{
     \institution{National University of Singapore}
 }
 \email{dcscts@nus.edu.sg}

\begin{abstract}
Fashion knowledge helps people to dress properly and addresses not only physiological needs of users, but also the demands of social activities and conventions. It usually involves three mutually related aspects of: occasion, person and clothing. However, there are few works focusing on extracting such knowledge, which will greatly benefit many downstream applications, such as fashion recommendation.
In this paper, we propose a novel method to automatically harvest fashion knowledge from social media. We unify three tasks of occasion, person and clothing discovery from multiple modalities of images, texts and metadata. For person detection and analysis, we use the off-the-shelf tools due to their flexibility and satisfactory performance.
For clothing recognition and occasion prediction, we unify the two tasks by using a contextualized fashion concept learning module, which captures the dependencies and correlations among different  fashion concepts. To alleviate the heavy burden of human annotations, we introduce a weak label modeling module which can effectively exploit machine-labeled data, a complementary of clean data. In experiments, we contribute a benchmark dataset and conduct extensive experiments from both quantitative and qualitative perspectives. The results demonstrate the effectiveness of our model in fashion concept prediction, and the usefulness of extracted knowledge with comprehensive analysis.
\end{abstract}



\maketitle
\section{Introduction} \label{introduction}

According to Statista\footnote{ https://www.statista.com/outlook/244/100/fashion/worldwide}, revenue in fashion market amounts to \$600 billion dollars in 2019, which demonstrates the great opportunities for various fashion related research and applications. Fashion knowledge plays a critical role in this area. It addresses not only physiological needs of users, but also the demands of social activities and conventions~\cite{song2017neurostylist,yu2018aesthetic,hidayati2018dress}. Take the fashion recommendation as an example, the recommender system should be capable of instructing a man to wear thick clothes in winter rationally with pants instead of shorts, and to dress suits in a formal conference.

\begin{figure}[htp]
	\centering
	\includegraphics[scale=0.4]{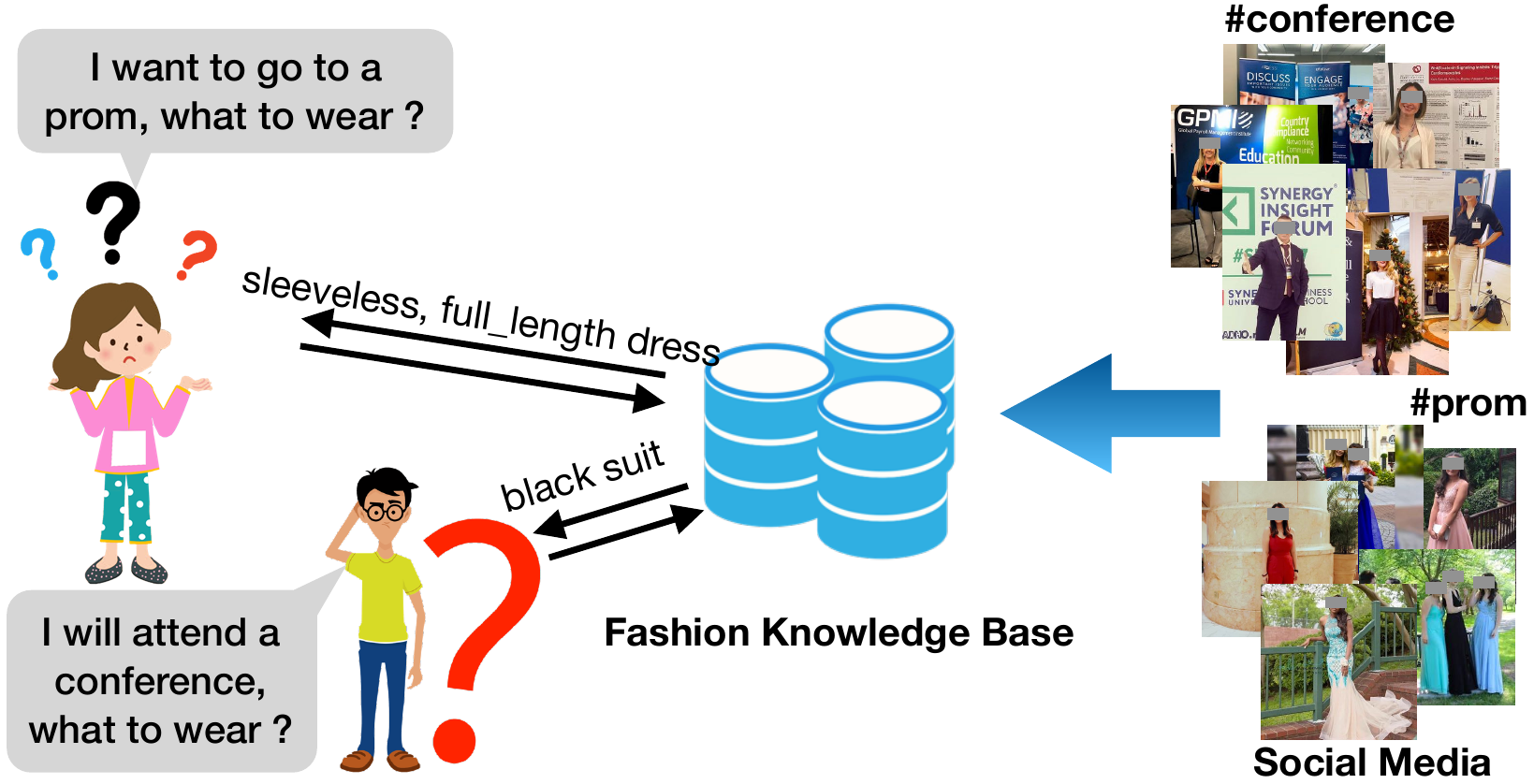}
		\vspace{-0.15in}
	\caption{An illustration of fashion knowledge extraction from social media. 
	}
	\label{Fig:illustration}
	\vspace{-0.15in}
\end{figure}

\begin{figure*}[tp]
	\centering
	\includegraphics[scale=0.45]{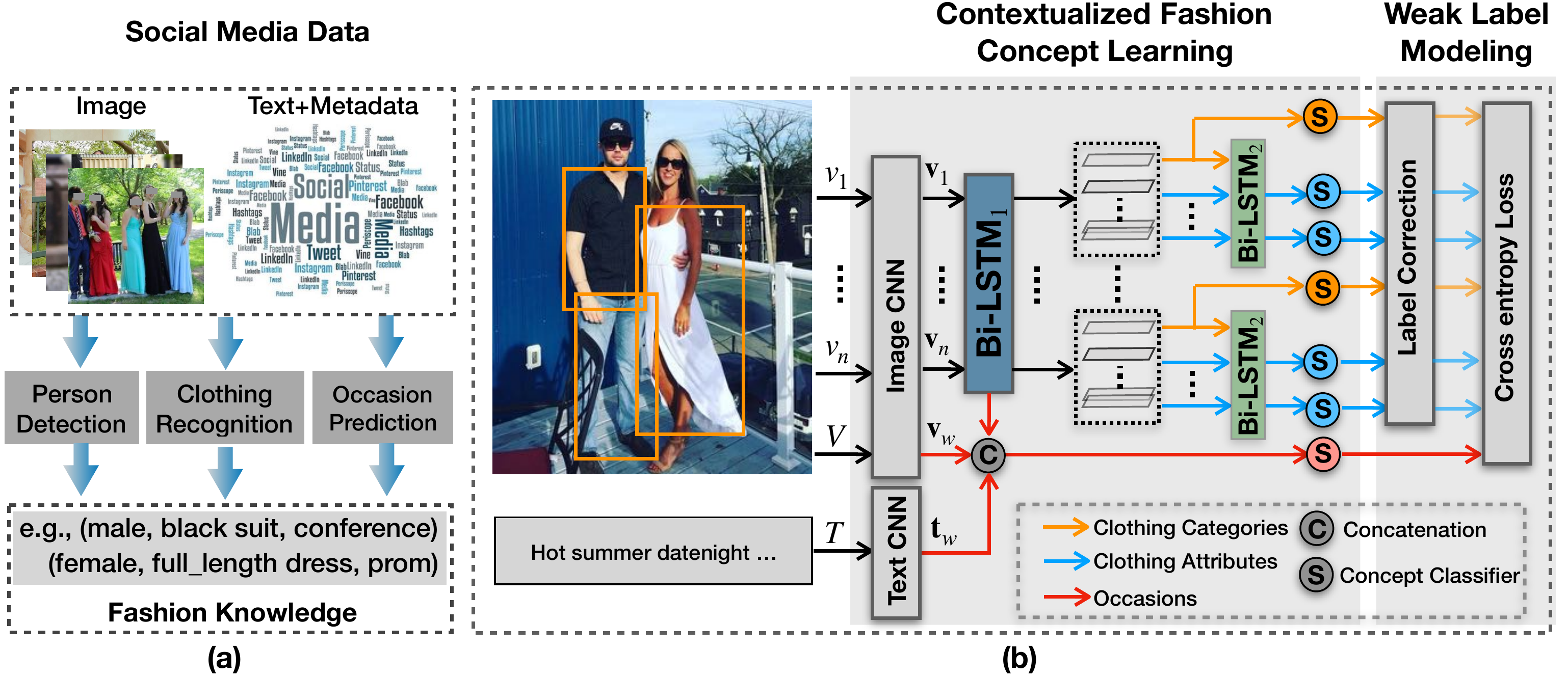}
	\vspace{-0.15in}
	\caption{(a) The overall framework of fashion knowledge extraction from social media, and (b) the pipeline of our proposed contextualized fashion concept learning model. Note that the extraction of person attributes (\textit{e.g.}, gender and age) is performed using an off-the-shelf tool for simplicity.
	}
	\vspace{-0.15in}
	\label{Fig:model}
\end{figure*}

Although some existing works focus on recognizing fashion concepts like clothing category and attributes \cite{huang2015cross,liu2016deepfashion,liao2018knowledge}, there are few studies at knowledge level in fashion domain, which usually involve three main aspects at the same time:  \textit{person}, \textit{clothing}, and \textit{occasion}. As illustrated in Figure \ref{Fig:illustration}, it would be better for a young girl to wear a sleeveless full length dress for a prom. To dress properly, people have to consider social conditions like occasion and personal identity, in addition to clothes. Clearly, there exist a large number of such patterns (\textit{e.g.}, dresscode or conventions) guiding people's daily fashion activities. But as we move forward, a key question arises: where and how can we collect such fashion knowledge? 

This paper proposes to automatically extract user-centric fashion knowledge from social media, such as Instagram, where massive user-generated multi-modal resources are uploaded every day from all over the world. It is a natural and appropriate source to extract fashion knowledge from general users, because (1) the images posted on social media usually contain the cues to various occasions such as the conference, wedding and travel \textit{etc.}, and also person identity information, such as the gender, age \textit{etc}, and (2) there is sufficient up-to-date data to perform the analysis. However, it is a non-trivial task due to the following challenges:

First, the extraction of fashion knowledge from social media content is highly dependant on the performance of fashion concept prediction which still remains unsatisfactory. This is because most of the visual images posted by various users on social media are taken in natural scenes, from which it is hard to detect the fashion concepts (\textit{e.g.}, clothing attributes and occasions). It is also more complex than the fashion product images typically with clean backgrounds, which most existing research studies focus on. Therefore, how to jointly detect the fashion concepts in the natural scene images for knowledge construction is a difficult but critical task.

Second, social media data lacks sufficient fashion concept labels which are crucial for fashion knowledge construction. The quality of automatically harvested fashion knowledge highly depends on semantic-level fashion concept learning. However, manually annotating a large amount of data is expensive and time-consuming. Existing datasets are mainly derived from e-commerce site and only focus on a specific set of cloth attributes, which cannot be used to detect the types of occasions or person identities. 

To address these challenges, we propose a novel method with two modules that jointly detect the fashion concepts using weakly labeled data. We propose a contextualized fashion concept learning module to effectively capture the dependencies and correlations among different fashion concepts.
To alleviate the label insufficiency problem, we enrich the learning procedure with a weak label modeling module that utilizes both the machine-labeled data and clean data. In particular, we incorporate a label transition matrix into this module to enable more robust noise control during the learning process. 
We then obtain a set of fashion concepts grounded with social media. 
Finally, through a statistical approach, we obtain the ultimate fashion knowledge. Through extensive evaluations and analyses, we demonstrate that the extracted knowledge is rational and is able to be applied to downstream applications.

The main contributions of this work are as follows:
\begin{itemize}
\item[$\bullet$] We propose a novel method for fashion knowledge extraction with the help of a contextualized fashion concept learning module, which is able to capture the dependencies among occasion, clothing categories and attributes.
\item[$\bullet$] We exploit machine labeled data with weak labels to enrich our learning model with a label correction module for noise control.
\item[$\bullet$] We contribute a benchmark dataset and conduct extensive experiments from both quantitative and qualitative perspectives to demonstrate the effectiveness of our model in fashion concept prediction and the usefulness of extracted knowledge. 
\end{itemize}

\section{Related Work} \label{related}
\noindent \textbf{Automatic Knowledge Extraction.}
In the past few years, researchers from the natural language processing (NLP), data mining, and computer vision communities have conducted extensive studies on automatic knowledge extraction and its applications \cite{KPRN, KGAT}. In the NLP community, several famous knowledge bases were curated such as YAGO \cite{fabian2007yago}, Freebase \cite{bollacker2008freebase}, WikiData \cite{vrandevcic2014wikidata}, DBpedia \cite{lehmann2015dbpedia}. These knowledge bases capture large amount of textual facts in the world, which are usually organized into triplets of the form (Subject, Predicate, Object). Most of the facts are about well-known people,  places,  and things, which were collected whether by crowd sourcing strategies or from large-scale semi-structured web knowledge bases. Nonetheless, all of them were curated only based on textual resources while neglecting the rich information existing in visual data. Thereafter, many efforts have been paid to extracting knowledge from visual data, such as NEIL \cite{Chen_2013_ICCV}, Visual Genome \cite{krishna2017visual} and VidVRD \cite{shang2017video}. 
Even though many researches targeted at extracting knowledge from both textual and visual data, few works aim to extract knowledge in vertical domains like fashion. \\
\noindent \textbf{Fashion Concept Prediction.}
Recently, fashion concept prediction has attracted increasing interests in various tasks such as clothing recognition \cite{chen2012describing,liu2016deepfashion}, retrieval \cite{liu2012street,huang2015cross,hadi2015buy,liao2018interpretable}, parsing \cite{yang2014clothing} and landmark detection \cite{liu2016deepfashion,wang2018attentive}. Earlier methods \cite{chen2012describing,liu2012street} mostly relied on handcrafted features (\textit{e.g.}, SIFT, HOG) to get good clothing representations. However, with the proliferation of deep learning in computer vision, many deep neural network models have been developed. In particular, Huang \textit{et al.} \cite{huang2015cross} developed a Dual Attribute-Aware Ranking Network (DARN) for clothing image retrieval. Liu \textit{et al.} \cite{liu2016deepfashion} proposed a branched neural network FashionNet which learns clothing features by jointly predicting clothing attributes and landmarks. Liao \textit{et al.} \cite{liao2018interpretable} introduced a novel data structure of EITree, which organizes the fashion concepts into multiple semantic levels and demonstrates good performance for both fashion image retrieval and clothing attributes prediction. However, most of the models are limited to the fashion concept level, while none of them further extended to fashion knowledge level. 
Moreover, data from social media lacks high-quality annotations, and weakly supervised methods are usually employed. Corbiere \textit{et al.} \cite{corbiere2017leveraging} learned a model from noisy datasets crawled from e-commerce websites without manual labelling, which demonstrates great generalization capability on DeepFashion \cite{liu2016deepfashion} dataset. However, it requires a lot of training data (1.3 million images in \cite{corbiere2017leveraging}), which was both time consuming and computationally intensive. In this paper, we also take advantage of weakly-labeled data to enhance our fashion concept prediction model. To counter the noise within the weak labels, we employ a weak label modeling approach, inspired by works on learning with noisy labels \cite{sukhbaatar2014training,tanaka2018joint}.

\section{Problem Formulation} \label{problem_formulation}
Our goal is to extract user-centric fashion knowledge from social media, such as Instagram, where massive user-centric multimodal resources are uploaded every day. We expect to obtain structured knowledge about \textit{what to wear for a specific occasion} to support downstream fashion applications. We first formally define the user-centric fashion knowledge as triplets of the form $\mathcal{K}=\{\mathcal{P}, \mathcal{C}, \mathcal{O}\}$, which consists of three aspects defined as follows:
\begin{itemize}
	\item \textbf{Person: } $\mathcal{P}$ refers to a set of person attributes, such as gender, age, body shape, etc. $\mathcal{P}$ should be able to describe a specific type of person, such as \textit{a young woman}.  
    \item \textbf{Clothing: } $\mathcal{C}$ refers to a set of clothing categories and attributes, such as skirt, a-line, red , etc. $\mathcal{C}$ should be able to describe a specific type of clothing, such as \textit{a red a-line skirt}. 
    \item \textbf{Occasion: } $\mathcal{O}$ refers to a set of occasions, such as conference and dating, and their affiliated metadata, such as location and time.
\end{itemize}
Given a set of user-generated posts $\mathcal{X}=\{\mathcal{V}, \mathcal{T}, \mathcal{M}\}$ consisting of images $\mathcal{V}$, texts $\mathcal{T}$, and metadata $\mathcal{M}$ (such as time, location) on social media, the problem is to develop a hybrid detection framework which is able to automatically extract the three aspects of fashion knowledge $\{\mathcal{P}, \mathcal{C}, \mathcal{O}\}$. 

Three sub-tasks that need to be tackled are: 1) person attributes detection, 2) clothing categories and attributes detection, and 3) occasion prediction. As existing person detection and analysis methods (such as \cite{redmon2018yolov3} and \cite{zhang2016joint}) have already achieved satisfactory performance, they can be utilized as off-the-shelf tools for our person information extraction. For simplicity, we only handle the last two sub-tasks in this work.

The main task of this work is to design a fashion concept learning framework as shown in Figure \ref{Fig:model} (b), which should jointly detect occasion and clothing categories and attributes from social media, the details of which will be presented in Section \ref{methodology}.


\noindent {\textbf{Dataset:}} Currently, there is no available training and evaluation dataset for the task of extracting \textit{user-centric fashion knowledge}.
Social media sites such as Instagram provide a huge amount of user generated contents of which a large portion involves people's dressing codes. There are many active fashion influencers who like to share their dressing styles and mix-and-match experience, which provide a valuable and up-to-date source for fashion knowledge extraction.
We thus crawl millions of posts from Instagram. Both automated and manual filtering are carried out sequentially to ensure data quality. Finally, we contribute a large and high quality dataset, named FashionKE, which consists of 80,629 eligible images. Each collected image is selected to be easily recognizable and diverse from the three aspects of: person, occasion, and clothing. We first leverage pre-trained object detection model \cite{redmon2018yolov3} to detect person body and face \cite{zhang2016joint}. Second, we filter out those images without any face or body. Moreover, we align the body and face in the same image and remove those images which do not have normal-sized body or face. 
Third, to ensure that the images are really user-generated -- but not advertisements or posters -- we manually check all the images and remove those that cannot reflect any occasion. 
The ontology of our dataset is shown in Figure~\ref{Fig:category_attr_tax} and the annotation of the dataset is described as follows:
\begin{figure}[tp]
	\centering
	\includegraphics[scale=0.56]{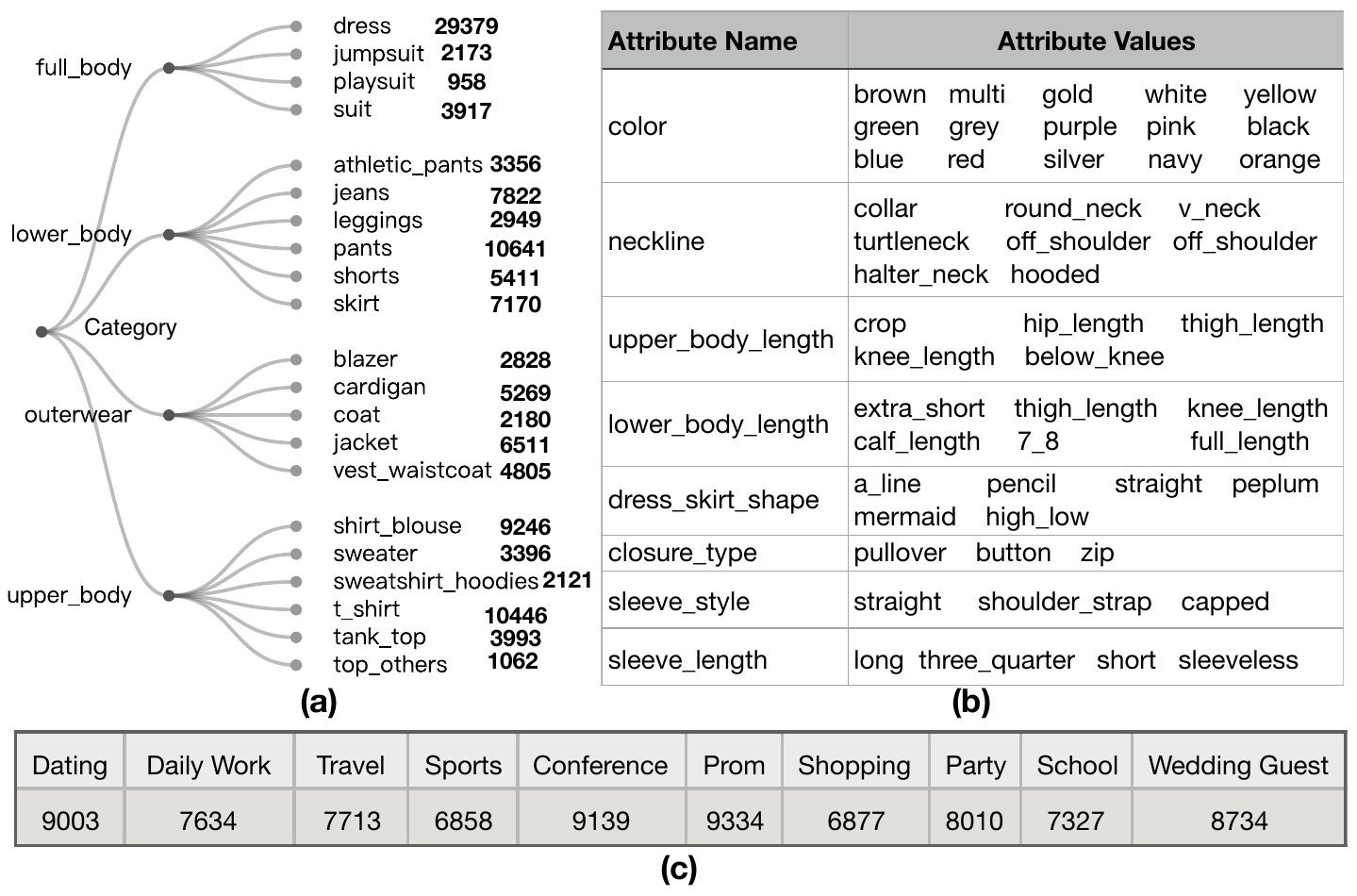}
			\vspace{-0.15in}
	\caption{The statistics of the FashionKE dataset, consisting of 21 categories, 8 types of clothing attributes, and 10 common types of occasions.}
	\vspace{-0.2in}
	\label{Fig:category_attr_tax}
\end{figure}

\noindent\textbf{Occasion annotation}. \textit{Occasion} is an important fashion concept that deeply affects people's decisions on dressing. With the help of fashion experts, we manually define 10 common types of occasion concepts as shown in Figure~\ref{Fig:category_attr_tax} (c). For each occasion, we curate a seed list of hashtags which are widely adopted and highly-correlated with that occasion. We then crawl Instagram posts using such hashtags and each post will have a potential occasion tag corresponding to its seed hashtag. We ask each annotator to give each post a binary label: whether such a post reveals that occasion or not by jointly considering the hashtags, post texts, and image content. It shows that such a process is efficient and effective. Finally, we obtain about 8,000 images for each occasion on average. \\
\noindent \textbf{Annotation of clothing categories and attributes}. This task is very important but time-consuming and expensive, since each image may have multiple persons and multiple sets of clothing. Compared with occasion annotation which requires only one label for each image, clothing annotation needs about 10 labels (categories and attributes of multiple clothes) for each image. To alleviate this issue, we adopt a two-stage annotation process: we first use a commercial fashion tagging tool\footnote{www.visenze.com} to automatically detect and tag the clothes, and then manually check and refine the results by human annotators. 
The statistic of the categories of all clothes is shown in Figure~\ref{Fig:category_attr_tax} (a). Note that only 30\% of images are carefully refined by human annotators. The rest of data are machine-labeled which are noisy. Therefore, how to exploit the machine-labeled noisy data is also one of our research questions. \\

\section{Our Approach} \label{methodology}
This paper proposes to develop a hybrid detection framework to extract user-centric fashion knowledge $\mathcal{K}=\{\mathcal{P}, \mathcal{C}, \mathcal{O}\}$ from social media. As mentioned before, we only focus on detecting clothing categories and attributes, and the occasions. The keys to tackling such a task are: 1) how to design a unified detection framework which is able to effectively capture the correlation among occasions, clothing categories and attributes; and 2) how to effectively utilize the machine labeled data to enhance the fashion concept learning. 

For the first question, we design {a contextualized fashion concept learning model, as shown in figure \ref{Fig:model}(b), in which two bidirectional recurrent neural networks are utilized to capture the dependencies and correlations among occasions, clothing attributes and categories}, which is presented in section \ref{Concept-learning}. For the second question, we introduce a weak label modeling module which estimates a label transition matrix for bridging the gap between weak labels and clean labels, as described in section \ref{weak-label}.

\subsection{Contextualized Fashion Concept Learning}\label{Concept-learning}
Given a post image ${V}$ and the affiliated text description $T$, we first detect a set of clothing regions $\{{v}_1, \cdots, {v}_i, \cdots,  {v}_M\}$ in the image ${V}$ by a clothing detection module. The goal is to predict the occasion label $\hat{y}_{o_V}\in\{y_{o_1}, \cdots, y_{o_{M_o}}\}$ of the posted image ${V}$ and the clothing category $\hat{y}_{c_{v_i}}\in\{y_{c_1}, \cdots, y_{c_{M_c}}\}$ and attributes $\hat{y}_{a_{v_i}}\in\{y_{a_1}, \cdots, y_{a_{M_a}}\}$ of each clothing region ${\{v\}}_{i=1}^M$. A simple solution is to directly cast it as three independent classification tasks. However, such a straightforward approach may result in sub-optimal performance as it ignores the relations between the occasion, clothing categories, and clothing attributes. For example, it is not likely for a woman to dress shorts or tanks to attend a prom. 
 
We develop a contextualized fashion concept learning framework to capture the correlations among the occasion, clothing categories, and clothing attributes, as shown in Figure~\ref{Fig:model} (b).\\
\noindent{\textbf{Category Representation: }}The first step is to learn the contextualized representations of clothing regions for category prediction and the whole image for occasion prediction. We use a bidirectional Long Short-Term Memory (Bi-LSTM) network to encode the dependence among all clothing regions. We first use a pre-trained convolutional neural network (CNN), such as ResNet~\cite{he2016deep} as our main feature extractor to extract the dense vector representation of the whole image ${V}$ as $\mathbf{v}_w\in\mathbb{R}^d$, and the vector representation of each clothing region $v_i\in\{v\}_{i=1}^M$ as $\mathbf{v}_i\in\mathbb{R}^d$. The final hidden representation for each clothing region is the concatenation of the hidden vectors in both directions:
\begin{equation}
\left\{
\begin{aligned}
\overrightarrow{\mathbf{h}}_{v_i} &= \overrightarrow{\mathrm{LSTM}_1}(\mathbf{v}_i, \overrightarrow{\mathbf{h}}_{\mathbf{v}_{i-1}})\\
\overleftarrow{\mathbf{h}}_{v_i} &= \overleftarrow{\mathrm{LSTM}_1}(\mathbf{v}_i, \overleftarrow{\mathbf{h}}_{v_{i+1}})\\
{\mathbf{h}}_{v_i} &=[\overrightarrow{\mathbf{h}}_{v_i}, \overleftarrow{\mathbf{h}}_{v_i}]
\end{aligned}
\right.
\end{equation}
where ${\mathbf{h}}_{v_i}\in{\mathbb{R}^{2d}}$. We then add a fully connected layer $F_c(\cdot)$ parameterized with a weight matrix $\mathbf{W}_c\in{\mathbb{R}^{2d\times d}}$ and a bias vector $\mathbf{b}_c\in{\mathbb{R}^{d}}$ to transform ${\mathbf{h}}_{v_i}$ as the final category representation of each clothing region $\mathbf{c}_{v_i}=\mathbf{W}_c^T \mathbf{h}_{v_i} + \mathbf{b}_c$.

\noindent{\textbf{Occasion Representation: }}To better represent the whole image $V$, we augment the CNN feature $\mathbf{v}_w$ with the feature of the post text description $\mathbf{t}_w$ and the final hidden state representation $\mathbf{h}_o = [\overrightarrow{\mathbf{h}}_{o}, \overleftarrow{\mathbf{h}}_{o}]\in{\mathbb{R}^{2d}}$ of Bi-LSTM in Eq. (1): 
\begin{equation}
\mathbf{v}'_w  =[\mathbf{v}_w, \mathbf{t}_w, \mathbf{h}_o]
\end{equation}
where $\mathbf{h}_o\in{\mathbb{R}^{2d}}$ encodes the inter-correlation of different clothing regions extracted from the whole image ${V}$. $\mathbf{t}_w\in\mathbb{R}^{d_t\in{\mathbb{R}^{d}}}$ denotes the vector representation of the affiliated text description $T$ which usually contains the evidence about the \textit{occasion}.  It is extracted by a TextCNN~\cite{kim2014convolutional}. Both $\mathbf{h}_o$ and $\mathbf{t}_w$ can effectively complement the CNN feature $\mathbf{v}_w$ of the whole image. A fully-connected layer $F_w(\cdot)$, parameterized with a weight matrix $\mathbf{W}_w\in{\mathbb{R}^{4d\times d}}$ and a bias vector $\mathbf{b}_w\in{\mathbb{R}^{d}}$, is added to transform the concatenated representation $\mathbf{v}'_w$ of the whole image to a $d$-dimensional occasion representation $\mathbf{o}_w=\mathbf{W}_w^T \mathbf{v}'_w+ \mathbf{b}_w$ for the whole image.

\noindent{\textbf{Attribute Representation: }}
Since each cloth has multiple different types of attributes, such as color, shape, sleeve length, \textit{etc.}, we introduce a multi-branch attribute prediction module, which consists of 
$K$ fully-connected layers $F_{a_i}(\cdot), i=1, \cdots, K$, parametrized with weight matrices $\mathbf{W}_{a_i}\in{\mathbb{R}^{2d\times d}}$ and bias vectors $\mathbf{b}_{a_i}\in{\mathbb{R}^{d}}$, to transform each clothing region representation $\mathbf{h}_{v_i}\in{\mathbb{R}^{2d}}$ into $K$ semantic representations $F_{a_k}(\mathbf{h}_{v_i}), k=1, \cdots, K$, for attribute prediction. Each branch corresponds to a type of clothing attribute. In this multi-branch structure, the visual representations from the lower-level layers are shared among all attributes. The neuron number in the output-layer of each branch equals to the number of corresponding attribute values. 

To capture the dependence among clothing attributes and categories, we introduce the second Bi-LSTM network. For each clothing region, we stack the outputs of $K$ branches $\{F_{a_k}(\mathbf{h}_{v_i})\}_{k=1}^K$ and the category representation $\mathbf{c}_{v_i}$ into a sequence of vectors as the inputs to the second Bi-LSTM. The final hidden representation for each attribute is the concatenation of the hidden vectors in both directions:
\begin{equation}
\left\{
\begin{aligned}
\overrightarrow{\mathbf{h}}^{v_i}_{a_k} &= \overrightarrow{\mathrm{LSTM}_2}(F_{a_k}(\mathbf{h}_{v_i}), \overrightarrow{\mathbf{h}}^{v_i}_{a_{k-1}})\\
\overleftarrow{\mathbf{h}}^{v_i}_{a_k} &= \overleftarrow{\mathrm{LSTM}_2}(F_{a_k}(\mathbf{h}_{v_i}),\overleftarrow{\mathbf{h}}^{v_i}_{a_{k+1}})\\
{\mathbf{h}}^{v_i}_{a_k} &=[\overrightarrow{\mathbf{h}}^{v_i}_{a_k}, \overleftarrow{\mathbf{h}}^{v_i}_{a_{k}}]
\end{aligned}
\right.
\end{equation}
where ${\mathbf{h}}^{v_i}_{a_k}\in{\mathbb{R}^{2d}}$ is a contextualized attribute representation which encodes the dependence among clothing attributes and categories. It is further transformed into a $d$-dimensional attribute representation $\mathbf{a}^{v_i}_{a_k}\in{\mathbb{R}^{d}}$ by a fully-connected layer.

After obtaining the occasion representation $\mathbf{o}_w$ of the whole image, the category representation $\mathbf{c}_{v_i}$ and attribute representation $\mathbf{a}^{v_i}_{a_k}$ of clothing regions, the prediction scores of occasions, clothing categories, and clothing attributes are obtained by multiple standard classifier layers (i.e., a linear function followed by a softmax layer), respectively. Cross-entropy loss is used to train the model. The training objective is to minimize the following loss function:
\begin{equation}\label{loss_clean}
\!L=\!L_o\!\left(V, y_{o_V},\Theta\right)+\!L_c\!\left(\{v_i\}, \{y_{c_{v_i}}\},\Theta\right)+\! L_a\!\left(\{v_i\}, \{y_{a_{v_i}}\},\Theta\right)
\end{equation}
where $L_o(\cdot)$, $L_c(\cdot)$, and $L_a(\cdot)$ denote the cross entropy losses of occasion, category, and attribute, respectively.

\subsection{Enhancing Fashion Concept Learning with Weak Label Modeling}\label{weak-label}
As aforementioned, only the occasion label is manually annotated for each image in our dataset. For the annotation of clothing categories and attributes, only a fraction of our dataset is provided with clean clothing category labels and attribute labels, while the rest of data is annotated by a fashion tagging tool. Such machine-labeled data is relatively cheap and easy-to-obtain, but the model would suffer from overfitting due to the label noise in training set. Therefore, how to jointly exploit machine-labeled data and the limited human-labeled data for model training is one of our main research focuses. This paper introduces a weak label modeling strategy to handle the label flip noise{ \cite{sukhbaatar2014training,tanaka2018joint}} in machine-labeled data. 

\noindent{\textbf{Weak Label Modeling}}. We first describe the process as a general setting. The goal is to learn a fashion concept learning model, as described in section \ref{Concept-learning}, from the machine-labeled data with weak label ${y'}\in\{1, 2, \cdots, N\}$. The true label is denoted as $y^*\in\{1, 2, \cdots, N\}$.
We assume that each weak label ${y'}$ depends only on the true label ${y^*}$ and not on the training sample, and further suppose that the weak labels are i.i.d. conditioned on the true labels. Then, we can represent the conditional noise model by a label transition matrix \cite{sukhbaatar2014training,tanaka2018joint} $\mathbf{Q}\in\mathbb{R}^{N\times N}$:
\begin{equation}
p({y}'=j|y^{\ast}=i)=q_{j,i}
\end{equation} 
 where $q_{j,i}$ is the element of label transition matrix $\mathbf{Q}$ at $(j, i)$. The probability of a data sample $\textbf{x}$ being labeled as a noisy label $j$ can be computed as:
\begin{equation}
p(\hat{y}'=j|\textbf{x}, \mathbf{Q}, \Theta) = \sum_i{q_{j,i}p(\hat{y}^{\ast}=i|\textbf{x}, \mathbf{\theta})} 
\end{equation}
Then, the prediction of weak label distribution is:
\begin{equation}\label{Eq.weakLabelDistribution}
p(\hat{y}'|\textbf{x}, \mathbf{Q}, \mathbf{\Theta})= \sum_i{p(\hat{y}'|\hat{y}^*=i)p(\hat{y}^*=i|\textbf{x}, \mathbf{\theta})} = \mathbf{Q} p(\hat{y}^{\ast}|\textbf{x}, \mathbf{\theta})
\end{equation}
where $p(y^{\ast}|\textbf{x}, \mathbf{\theta})$ is used to estimate the true label of testing sample, while $p(\hat{y}'|\textbf{x}, \mathbf{Q}, \mathbf{\Theta})$ is used for training with a standard cross-entropy loss. 

In summary, the basic idea is to add a \textit{label correction} layer with an estimated label transition matrix $\mathbf{Q}$ after the prediction layer of our network framework, as shown in figure \ref{Fig:model}(b), which adapts the prediction to match the weak label distribution.

\noindent{\textbf{Estimation of label Transition Matrix: }} How to effectively estimate the laebl transition matrix $\mathbf{Q}$ is critical to our weak label modeling module. In this work, we implement it as a linear layer to be jointly optimized during training. Since a fraction of our dataset is corrected by human annotators from machine-labeled data, we can first estimate a label transition matrix using the human-corrected labels and the weak labels in the clean part, which is further utilized as an initialization for the linear layer.

\noindent{\textbf{Learning: }} We split the training data into two sets: $\mathcal{X}^*$ and $\mathcal{X}'$ where $\mathcal{X}^*$ denotes the part of our training data with clean labels $\mathcal{Y}^*$, and $\mathcal{X}'$ denotes the rest of our training data with weak labels $\mathcal{Y}'$. Our final objective is to minimize the following fused cross-entropy loss function:
\begin{equation}
L=L^*(\mathcal{X}^*, \mathcal{Y}^*, \Theta^*) + \beta L'(\mathcal{X}', \mathcal{Y}', \mathbf{Q},\Theta')
\end{equation}
where $L^*(\cdot)$ is the loss function defined on clean data, as shown in Eq. (\ref{loss_clean}), and $L'(\cdot)$ is the loss function defined on machine-labeled data based on weak label prediction in Eq. (\ref{Eq.weakLabelDistribution}). $\beta$ is a trade-off hyperparameter.
\section{Experiments} \label{experiments}
To verify the effectiveness of fashion knowledge extraction, we conduct a series of experiments from both the quantitative and qualitative perspectives. We first compare it with three methods for fashion concept prediction on our new benchmark, and then analyze the fashion knowledge based on the extracted concepts. Particularly, we are interested in the following questions: \\
(1) \textbf{RQ1}: Does our fashion knowledge extraction model perform well in the preliminary step of concept prediction? \\
(2) \textbf{RQ2}: Why does our method achieve superior performance? \\
(3) \textbf{RQ3}: Whether the extracted fashion knowledge is reasonable and important to downstream applications? 
\subsection{Experimental Settings}
\textbf{Experimental Setup.} The dataset is constructed in a semi-supervised manner with manual correction (Section \ref{problem_formulation}). For evaluation, we split the dataset into two parts: 90\% for training (70\% machine-labeled data and 20\% clean data) and 10\% for testing. Note that all the testing data is randomly selected from the clean part. The evaluation metric is the standard accuracy.
\\
\textbf{Implementation Details.} For visual representations, we use the pretrained ResNet-18 \cite{he2016deep} as the feature extractor, which outputs a 512-D dense vector representation. As for the textual information, we utilize the pretrained 300-D word embedding (\textit{i.e.}, Glove \cite{pennington2014glove}) followed by a text CNN architecture \cite{kim2014convolutional}, which consists of a single channel, four kernels with sizes of \{2,3,4,5\}, and a max pooling layer, where each kernel has 32 feature maps and uses the rectified linear unit (ReLU) as the activation function. Finally, we obtain a 128-D vector for text representation. The hidden state size of two Bi-LSTM networks is set as 512. 
In terms of the order of input sequence of the first Bi-LSTM, we sort the clothing regions by their spatial positions (\textit{i.e.}, from left to right, top to bottom). For the second Bi-LSTM, we keep the order of attributes classifiers the same among all training samples. For weak label modeling, we implement the noise transition matrix with a fully-connected layer \cite{sukhbaatar2014training}. We initialize the label transition matrix with a statistical estimation on the part of training data with human-corrected labels. In particular, for each element $q_{j,i}$ of $\textbf{Q}\in{\mathbb{R}^{N\times{N}}}$ (Section \ref{weak-label}), we count the number of labels whose ground-truth are $i$ while their predictions are label $j$ by the tagging tool, and then normalize the estimated label transition matrix along each column. We empirically set the trade-off hyperparameter $\beta$ as 0.5 throughout the experiments.

Our model is implemented with the PyTorch framework. For optimization, we employ the stochastic gradient descent (SGD) \cite{bottou1991stochastic} with the momentum factor of 0.9. We set the initial learning rate as 0.001 for the text CNN and image CNN,  and $10^{-5}$ for the linear layer of label transition matrix. The learning rate drops by 10 after every 4 epochs. The performance of the model on the testing set is reported until convergence. 
\\
\textbf{Baseline Methods.} Since fashion knowledge extraction is a relatively new problem and there are few specific methods for solving it, we choose the following three state-of-the-art baselines that tackle one or more subtasks of fashion concept (\textit{i.e.}, occasion, clothing category and attributes) prediction. The results of predicted concepts play an important role in fashion knowledge analysis. 
\begin{itemize}
    \item \textbf{DARN} \cite{huang2015cross} adopted an attribute-regularized two-stream CNN for cross domain fashion retrieval with a multi-branch fashion concept predictor. We only keep one stream of DARN for our task. 
    \item \textbf{FashionNet} \cite{liu2016deepfashion} is a state-of-the-art model for both clothing landmark detection, and clothing category and attributes prediction, which demonstrates compelling performance in clothing category classification and attribute prediction. We remove the clothing landmark prediction branch in {FashionNet} in our experiments.
    \item \textbf{EITree} \cite{liao2018interpretable} is a state-of-the-art model aiming at multimodal retrieval for fashion products by leveraging a special label structure of EI tree. 
\end{itemize}

\subsection{Fashion Concept Prediction (RQ1)} \label{model_comparison}

Since there is no occasion classification module in all of the three baselines, we add an additional branch of occasion classifier into these baselines. For fair comparisons, we also remove the textual inputs because the three baselines are not designed to handle textual information. 
Table \ref{Tbl:context_comparison} shows the accuracy of predicting fashion occasion, category and attributes. We have the following observations:

First, our method outperforms all the baselines on all of the three tasks. This is mainly because: 1) our model takes advantage of machine-labeled data while suppresses the inherent label noise through a weak label modeling module, thus obtaining additional creditable supervisions; and 2) we consider the dependencies and correlations among the occasion, clothing category and attributes, which provide additional discriminating capability for fashion concept prediction. Such dependencies and correlations among multiple fashion concepts implicitly demonstrate the existence of fashion knowledge and its positive impacts on related applications.

Second, our method has been further improved by utilizing textual information, especially on occasion classification. This is because the short texts affiliated with social media posts usually contain rich occasion-aware descriptions, which are important for fashion knowledge extraction.

\subsection{Ablation Study (\textbf{RQ2})} \label{evaluate_submodules}
\subsubsection{Effect of Contextualized Fashion Concept Learning}
To evaluate the effect of the proposed contextualized fashion concept learning module, we further compare it with several variants listed below. In addition, we conduct experiments using only the clean data to get rid of the intervention of weak label modeling module, which is the concern of next section. We also remove the textual descriptions because it is modeled by another textCNN module.

\textbf{Base: } We remove the two Bi-LSTM modules without considering any concept dependencies and correlations, leading to a basic version of our model.

\textbf{$\textbf{Bi-LSTM}_1$: } We only keep the first Bi-LSTM network used to encode the dependency among co-occurring clothing regions.

\textbf{$\textbf{Bi-LSTM}_2$: } We only keep the second Bi-LSTM network used to capture the dependencies and correlations among clothing attributes and category.

\textbf{Final: } The proposed contextualized fashion concept learning model with two Bi-LSTM modules.

{ We can have the following observations from the results presented in Table \ref{Tbl:label_fusion_comparison}:

First,  Bi-$\mathrm{LSTM_1}$ outperforms the Base method on all three tasks, especially on category and occasion classification. This is due to two reasons: 1) Bi-$\mathrm{LSTM_1}$ improves the prediction of clothing category by modeling the dependencies among co-occurred clothing regions. It is reasonable, since the visual context of clothing regions in the same image is captured in this way, which results in contextualized representation of clothing regions. 2) It improves the prediction of occasions because we augment the CNN representation of occasion with the final hidden state of Bi-$\mathrm{LSTM_1}$. It makes sense since the final hidden state encodes the contextualized clothing information which complements the occasion representation significantly.

Second, Bi-$\mathrm{LSTM_2}$ again achieves better performance than the Base model, since it models the dependencies among different attribute representations and category representation.

Third, when using both the Bi-$\mathrm{LSTM}$ in the Final model, the performances of three predictions are all significantly improved. This demonstrates the necessity of employing both Bi-$\mathrm{LSTM_1}$ and Bi-$\mathrm{LSTM_2}$ to achieve mutual enhancement.
}
\begin{table}[tp]
    \centering
    \footnotesize
    \caption{Overall Performance.}
    \vspace{-0.16in}
    \label{Tbl:context_comparison}
    \begin{tabular}{ c|c|c|c } 
         \hline
         setting & occasion & category & attributes \\
         \hline
         \hline
         \textbf{DARN} \cite{huang2015cross} & 41.56\% & 68.04\% & 66.01\% \\
         \hline
         \textbf{FashionNet} \cite{liu2016deepfashion} & 41.53\% & 67.33\% & 65.6\% \\
         \hline
         \textbf{EITree} \cite{liao2018interpretable} & 39.64\% & 68.28\% & 63.95\% \\
         \hline
         \hline
         \textbf{our method w/o text} & 42.61\% & 73.6\% & 69.4\% \\ 
         \hline
         \textbf{our method} & \textbf{47.88\%} & \textbf{73.95\%} & \textbf{69.59\%} \\
         \hline
    \end{tabular}
    \vspace{-0.2in}
\end{table}
\begin{table}[htp]
\centering
\footnotesize
\caption{The performance comparison regarding two Bi-LSTM modules. Experiments are conducted on clean data.}
\vspace{-0.15in}
\label{Tbl:label_fusion_comparison}
\begin{tabular}{ c|c|c|c } 
 \hline
 setting & occasion & category & attributes \\
 \hline
 \hline
 \textbf{Base}  & 38.86\% & 69.89\% & 67.35\% \\
 \hline
 \textbf{$\textbf{Bi-LSTM}_1$} & 39.85\% & 70.82\% & 67.79\% \\ 
 \hline
 \textbf{$\textbf{Bi-LSTM}_2$} & 38.45\% & 71.31\% & 67.57\% \\ 
 \hline
 \textbf{Final} & \textbf{40.06\%} & \textbf{71.35\%} & \textbf{67.82\%} \\
 \hline
\end{tabular}
\vspace{-0.15in}
\end{table}

\subsubsection{Effect of Weak Label Modeling}
\begin{figure}[tb]
	\centering
	\vspace{-0.12in}
	\includegraphics[scale=0.45]{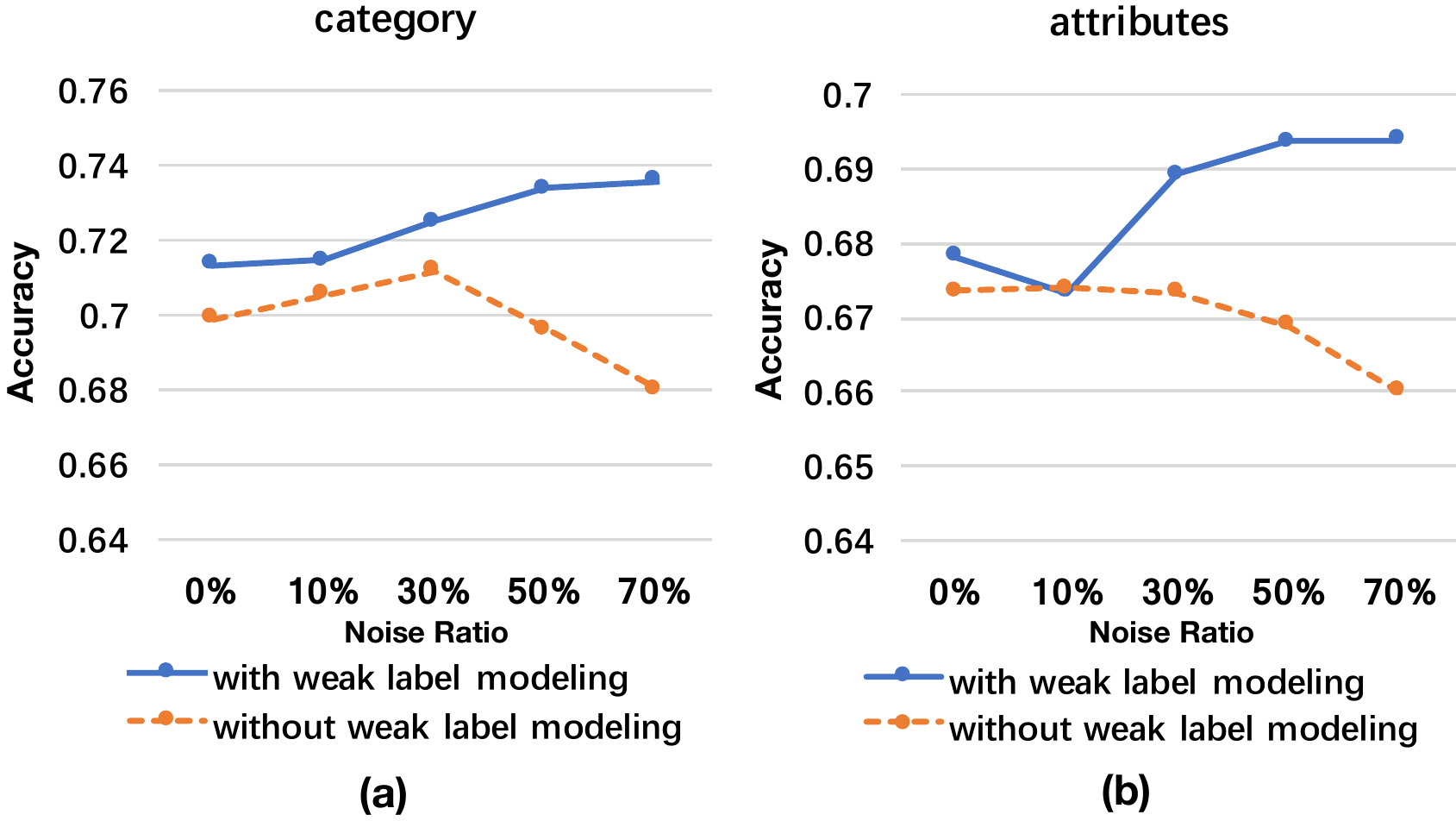}
		\vspace{-0.15in}
	\caption{Performance comparison with different ratios of machine-labeled data with weak labels.}
	\label{Fig:noise_comparison}
	\vspace{-0.15in}
\end{figure}

{ To verify the utility and robustness of our approach to learning with weak labels, we conduct experiments to compare our model's performance with different weak data ratios. As illustrated in Figure \ref{Fig:noise_comparison}, we gradually increase the weak data ratio along the x-axis to compare the performance of our method with/without weak label modeling module in Section \ref{weak-label}. When using the introduced weak label modeling module, we can observe a clear performance improvement with the increasing weak data ratio. However, if we remove the weak label modeling module, the performance of category prediction would first gradually improve and then degrade rapidly when the ratio is large than 30\%. This is because the flip label noise in machine-labeled data has dominated the optimization when the weak data ratio is beyond 30\%. It indicates the effectiveness of the weak label modeling module which performs label correction by a label transition matrix to match the distribution of weak labels.
With such a weak label modeling module, we can update our model with massive training data with weak labels.}

\subsection{Obtaining and Analyzing Fashion Knowledge (\textbf{RQ3})}
Given the predicted fashion concepts, the triplet form of fashion knowledge (occasion, person, clothing) will constitute a piece of knowledge that provides guidance for people's dressing in certain occasions. However, the fashion knowledge is subjective even if in the same conventions. Hence it is difficult to formulate a piece of convincing fashion knowledge from a single occurrence. Therefore, in this section, we discuss how to obtain useful fashion knowledge based on concepts from statistical perspective. The basic idea is that a piece of fashion knowledge is useful when it is widely adopted. As far as we know, this is the first work focusing on fashion knowledge rather than the concepts.

\subsubsection{From Fashion Concepts to Knowledge}
\begin{figure}[t]
	\centering
	\includegraphics[scale=0.58]{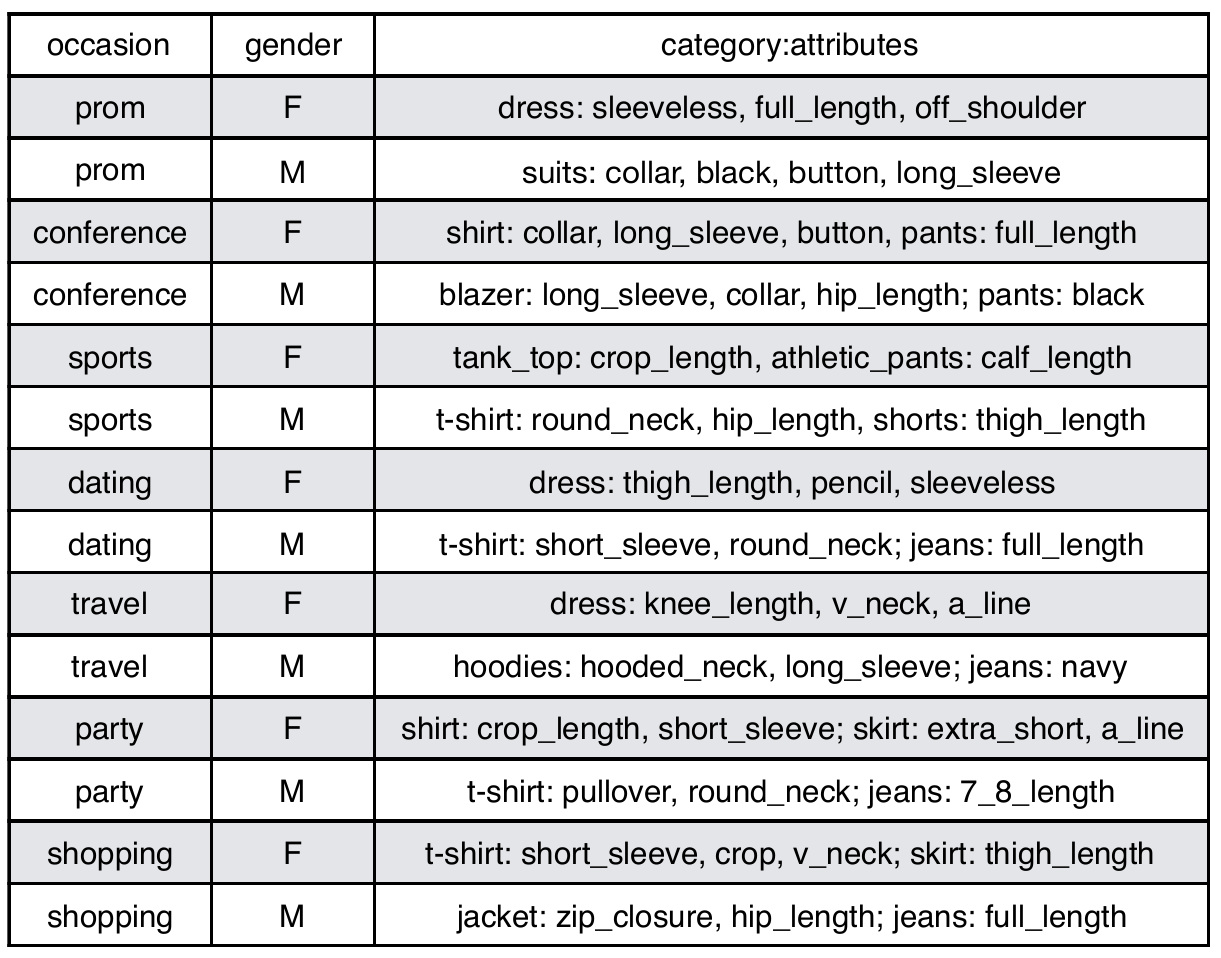}
		\vspace{-0.1in}
	\caption{Illustration of some pieces of fashion knowledge we obtained with high popularity.}
	\label{Fig:fashion_knowledge_generation}
	\vspace{-0.12in}
\end{figure}
\begin{figure}[tbh]
	\centering
	\includegraphics[scale=0.5]{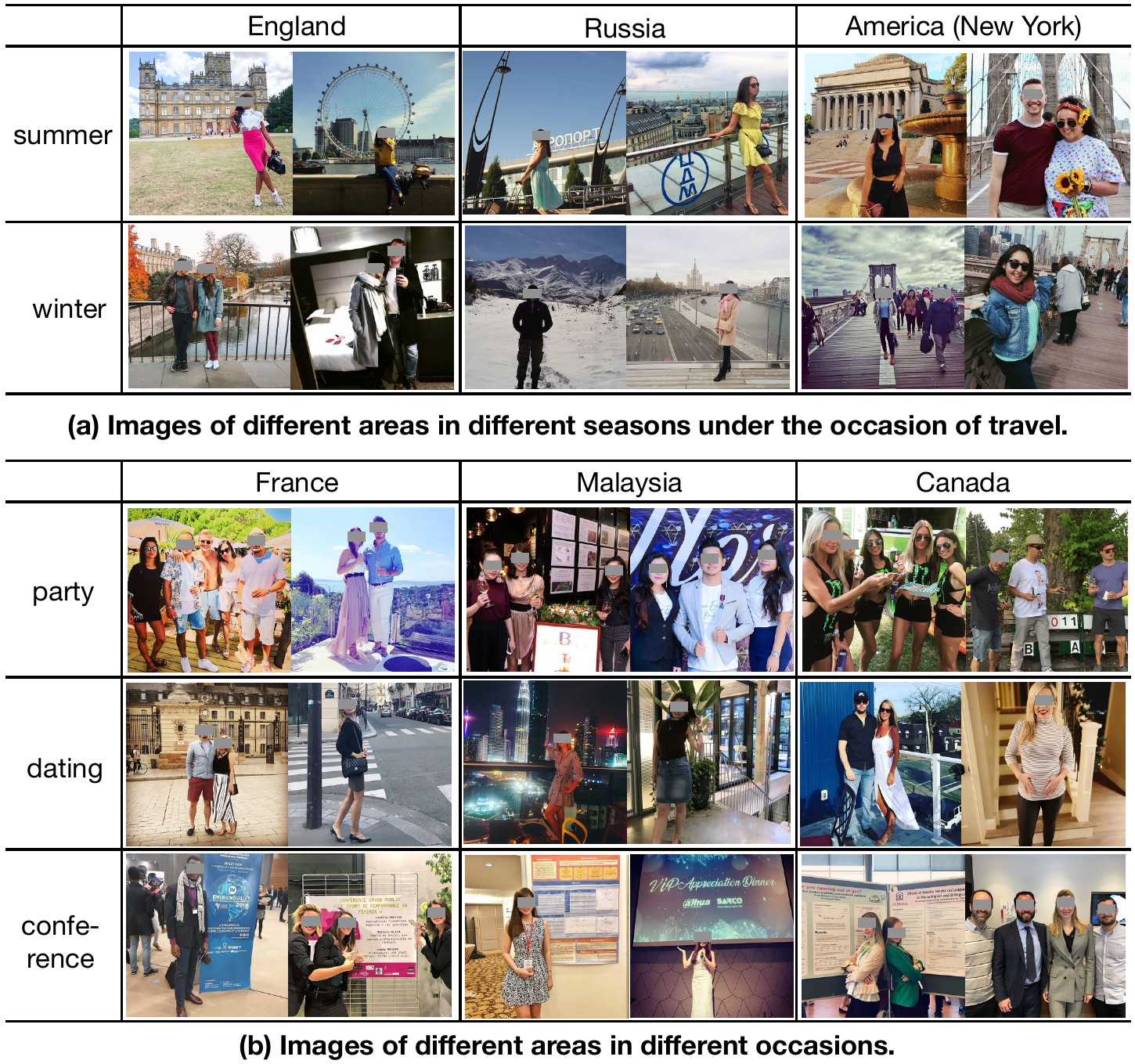}
		\vspace{-0.12in}
	\caption{Some exemplar images in different occasions. The top two rows (a) show the travel occasion in three different areas. The bottom part (b) demonstrates three occasions in three different areas.}
	\vspace{-0.2cm}
	\label{Fig:example_images}
	\vspace{-0.15in}
\end{figure}

\begin{figure*}[tp]
	\centering
	\includegraphics[scale=0.6]{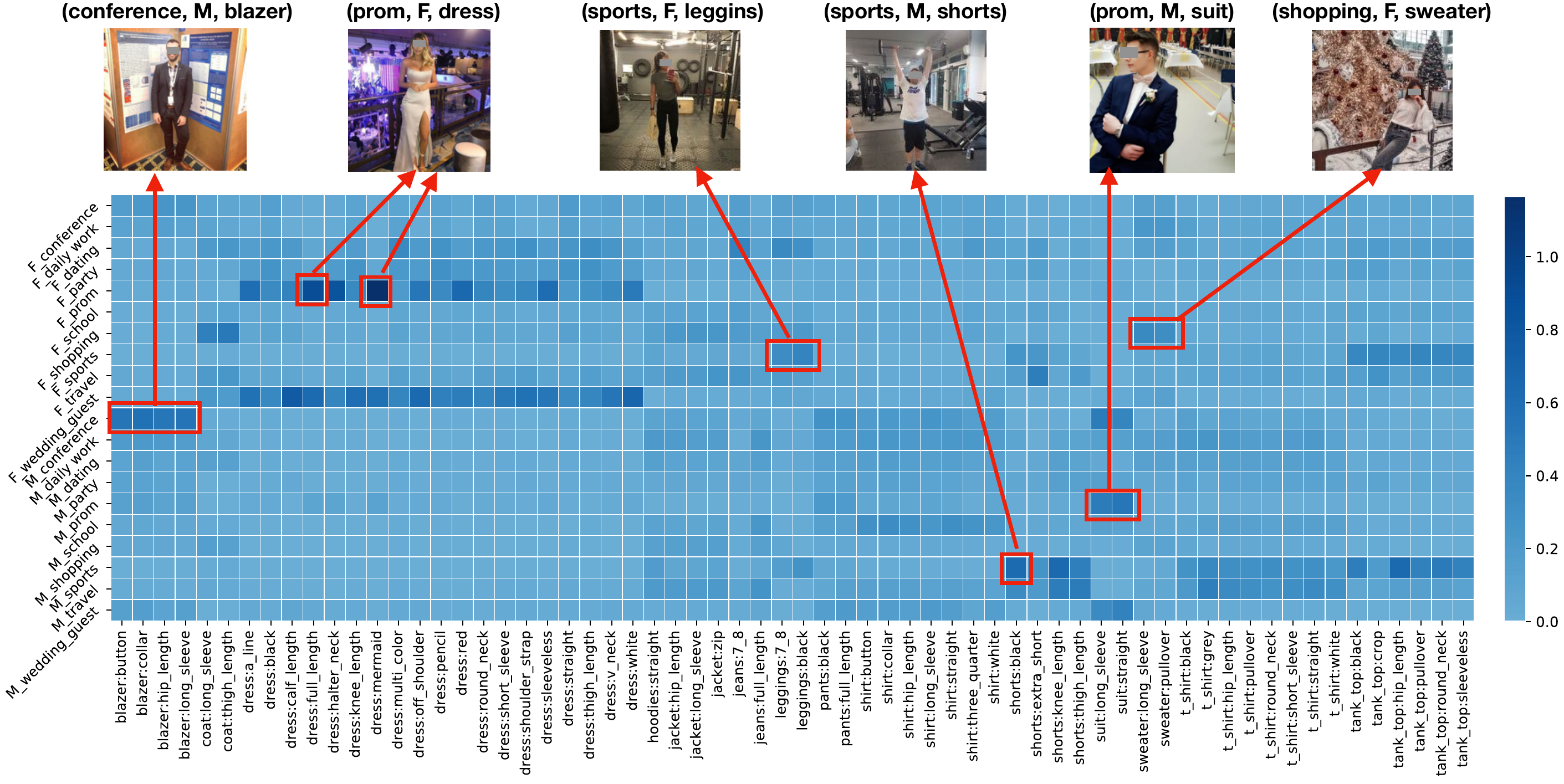}
		\vspace{-0.15in}
	\caption{The correlation between occasions and popular clothing. The horizontal axis demonstrates various popular clothes. The vertical axis shows the occasions for male (M-) and female (F-). The darker color means there are more images satisfying the condition of triplet at that point. The result is based on our model's prediction of testing data. 
	}
	\label{Fig:fashion_knowledge_analysis}
\end{figure*}
\begin{figure}[tb]
	\centering
	\includegraphics[scale=0.3]{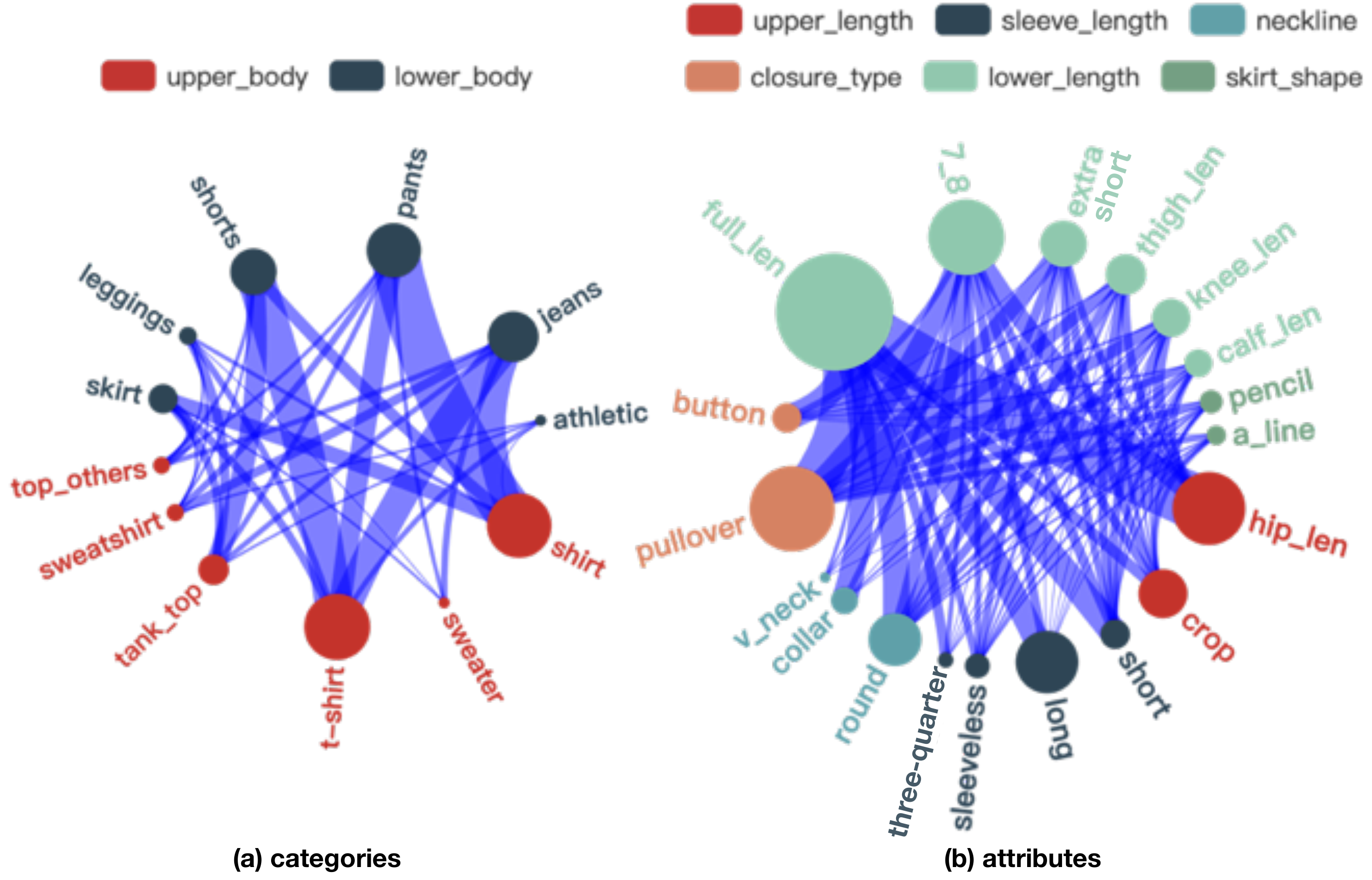}
		\vspace{-0.13in}
	\caption{The cross-category matching relationships between categories (a) and attributes (b). The size of each node shows the number of clothing and the strength of edges represents the number of connections. 
	}
	\label{Fig:cross_clothes_matching}
	\vspace{-0.2in}
\end{figure}
We employ a three-step process to summarize the most popular fashion knowledge from our extracted fashion concepts. First, we distinguish different clothes by their attributes. For example, two clothes are treated as the same only if they have the same category and attributes of interest. 
For efficiency, we use at most three attributes for two clothes' comparison. Second, we obtain the popular combinations of upper body clothes and lower body clothes, \textit{a.k.a}, outfit, by counting the combinations for upper body clothes and lower body clothes. Third, we count all the combinations of occasion, gender, and clothing (outfits), which are the triplets we defined in Section \ref{problem_formulation}. Finally we sort these triplets by their occurrence in descending order, and those triplets with higher frequencies are kept as useful fashion knowledge. Figure~\ref{Fig:fashion_knowledge_generation} shows some representative pieces of knowledge in different occasions.

\subsubsection{Fashion Knowledge Analysis} 
Figure~\ref{Fig:example_images} shows some exemplar images in different occasions, areas and seasons. We can see that images within the same occasion and location or season present some commondition. For example, in the \textit{conference} occasion, people all dress formally with either \textit{dress}, \textit{suit}, or \textit{blazer}. While in the \textit{travel} occasion, even though people wear different clothing, most of them are in casual style. Thus we can see that it is rational and reliable to extract the common dressing patterns under certain occasions. What's more, a piece of widely accepted fashion knowledge is highly probable to be grounded by a large amount of images, which have extensive commonness among them. 

On the other hand, different pieces of knowledge present rich and diverse information in terms of occasion, person, and clothing. 
First, occasion hugely influences the clothing distribution. For example, \textit{dress} is popular in \textit{wedding\_guest} and \textit{prom} occasions, \textit{blazer} is popular in \textit{conference} occasion, \textit{t\_shirt} and \textit{tank\_top} are popular in \textit{sports} and \textit{travel} occasions. However \textit{shorts} scarcely appear in the \textit{wedding\_guest} and \textit{prom} occasions, and \textit{dress} rarely appears in \textit{sports} occasion. This observation is in harmony with common sense and demonstrates that our extracted knowledge captures the correlation between occasions and clothing. Second, people with different genders have distinctive dressing styles. The top left part of Figure~\ref{Fig:fashion_knowledge_analysis}  indicates that females are more likely to wear \textit{dress} and the bottom right of Figure~\ref{Fig:fashion_knowledge_analysis} illustrates that males are more likely to wear \textit{t-shirt} and \textit{pants}. This phenomenon seamlessly verifies our claim in the very beginning of this paper that what to wear and how to wear are hugely affected by human identity.

Interestingly, we can also discover some insightful points by fine-grained comparisons. For example, in terms of the dresses in occasion \textit{F\_prom} and \textit{F\_wedding\_guest}, the attributes \textit{full\_length}, \textit{sleeveless}, and \textit{a\_line} are most popular, while \textit{thigh\_length} and \textit{pencil} are less popular. It makes sense that in a formal occasion like \textit{prom}, a \textit{dress} of \textit{full\_length}, \textit{sleeveless}, and \textit{a\_line} are much more formal than that of \text{thigh\_length} and \textit{pencil}. 

\textbf{Cross Category Matching}. Without considering the occasions, the cross category matching reveals the relationship among clothes themselves, which will benefit many downstream applications. For example, in fashion recommendation \cite{song2018neural,yang2018transnfcm,yang2019interpretable}, such cross-category matching rules can guide the model to recommend clothes that are functionally compatible with the given clothes. 
We illustrate the matching popularity between \textit{upper\_body} clothes and \textit{lower\_body} clothes in Figure \ref{Fig:cross_clothes_matching} (a). For example, \textit{shirts} are more likely to match with \textit{pants} other than \textit{shorts}, and \textit{tank\_top} is linked to \text{shorts} with the highest weight among all the connections. Similar matching patterns also exist between attributes. As illustrated in Figure \ref{Fig:cross_clothes_matching} (b), \textit{long\_sleeves}'s connection with \textit{full\_length} has the highest strength but its connections with \textit{thigh\_length}, \textit{extra\_short}, \textit{pencil} are much weaker. 

\section{conclusion}
In this paper, we explored a new task of automatically extracting fashion knowledge from social media. To build an effective model for fashion concept prediction, we designed a contextualized fashion concept learning model and enhanced it with weak label modeling. And more importantly, the analysis of the extracted knowledge verifies our hypothesis that fashion are affected by three main aspects of person, occasion, and clothing.

There are several research directions that can be conducted in the future: 1) The use of the extracted fashion knowledge into various downstream applications such as fashion recommendation. 2) The extraction of more fine-grained knowledge. In the future, we will try to apply the proposed visual concept learning method to enhance the general visual retrieval tasks, such as cross-modal retrieval~\cite{hong2017coherent} and person retrieval~\cite{yang2017person,yang2017enhancing}.
\section*{acknowledgement}
This research is part of NExT++ project, which is supported by the National Research Foundation, Prime Minister's Office, Singapore under its IRC@SG Funding Initiative.

\bibliographystyle{ACM-Reference-Format}
\bibliography{mybib}

\end{document}